\documentclass[runningheads]{llncs}
\usepackage[T1]{fontenc}
\usepackage{pifont}
\usepackage{amsfonts}
\usepackage{multirow}
\usepackage{booktabs}
\usepackage{makecell}
\usepackage{graphicx}
\usepackage{xspace}
\usepackage{xcolor}
\usepackage{marvosym}

\begin{document}
%
\title{TP-DRSeg: Improving Diabetic Retinopathy Lesion Segmentation with Explicit Text-Prompts Assisted SAM}
\titlerunning{TP-DRSeg}
%
\author{Wenxue Li\inst{1,2} \and Xinyu Xiong \inst{3}
\and Peng Xia\inst{1,2}  
\and Lie Ju\inst{1,2}\textsuperscript{(\Letter)}
\and Zongyuan Ge \inst{1,2}\textsuperscript{(\Letter)}
}
\authorrunning{W. Li et al.}
\institute{
$^1$Monash-Airdoc Research, Monash University\\ 
$^2$Monash Medical AI Group, Monash University\\
$^3$School of Computer Science and Engineering, Sun Yat-sen University \\
\email{wxli408@gmail.com, \{Lie.Ju1, zongyuan.ge\}@monash.edu}}
\maketitle              
\begin{abstract}
Recent advances in large foundation models, such as the Segment Anything Model (SAM), have demonstrated considerable promise across various tasks. Despite their progress, these models still encounter challenges in specialized medical image analysis, especially in recognizing subtle inter-class differences in Diabetic Retinopathy (DR) lesion segmentation. In this paper, we propose a novel framework that customizes SAM for text-prompted DR lesion segmentation, termed TP-DRSeg. Our core idea involves exploiting language cues to inject medical prior knowledge into the vision-only segmentation network, thereby combining the advantages of different foundation models and enhancing the credibility of segmentation. Specifically, to unleash the potential of vision-language models in the recognition of medical concepts, we propose an explicit prior encoder that transfers implicit medical concepts into explicit prior knowledge, providing explainable clues to excavate low-level features associated with lesions. Furthermore, we design a prior-aligned injector to inject explicit priors into the segmentation process, which can facilitate knowledge sharing across multi-modality features and allow our framework to be trained in a parameter-efficient fashion. Experimental results demonstrate the superiority of our framework over other traditional models and foundation model variants. The code implementations are accessible at https://github.com/wxliii/TP-DRSeg.

\end{abstract}

\section{Introduction}

\begin{figure*}[t]
\includegraphics[width=\textwidth]{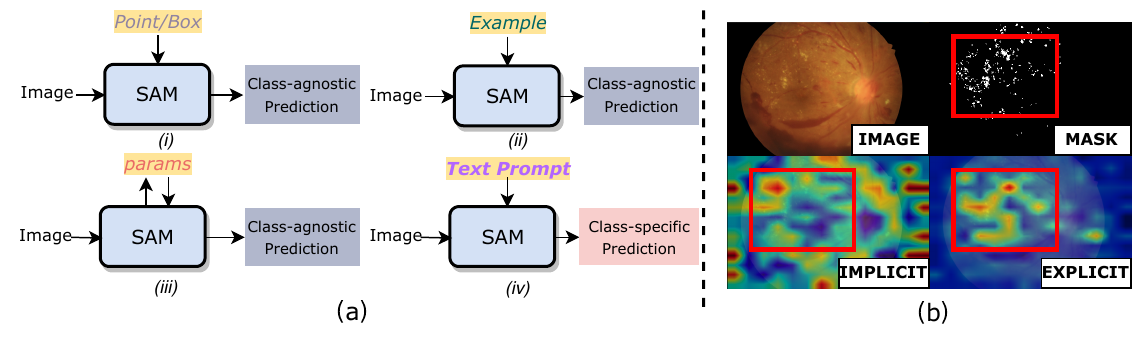}
\caption{
(a) Comparison of existing methods and our proposed method. (b) Class activation maps generate by CLIP~\cite{clip}. Bottom left utilizes text embedding from \textit{implicit class name} (hard
exudate). Bottom right employs text embedding from \textit{explicit description of lesions} (yellowish-white deposits). 
} 
\label{fig:title}
\end{figure*}

Diabetic retinopathy (DR) is a leading cause of visual impairment, becoming one of the world’s most serious health challenges. Automatic DR lesion segmentation is an important task in color fundus image analysis. Effective monitoring of certain lesions, including microaneurysms (MAs), hemorrhages (HEs), hard exudates (EXs), and soft exudates (SEs), provides vital assistance for ophthalmologists and significantly boosts early diagnostic accuracy and efficiency. 

Existing DR segmentation networks~\cite{NC19_LSEG,NC23_CLCNet,CIS22_CARNet,BSPC24_MSF,xia2024generalizing,hu2024ophnet} have achieved certain promising results by designing exquisite attention mechanisms to comprehend the visual features provided by the encoder. However, they still face two main challenges. On the one hand, with small vision backbones (such as ResNet-34) and limited training data available, these methods typically require a lengthy training process~\cite{BSPC24_MSF} to learn valuable representations, which is time-consuming and prone to over-fitting. On the other hand, the subtle inter-class differences pose challenges in accurate lesion classification. Existing trials only focus on vision supervision, lacking the guidance of specialized domain knowledge~\cite{xie2021survey}.  

Recent advances in foundational vision models have attracted considerable interest, such as the Segment Anything Model (SAM)~\cite{SAM}, showcasing fantastic capacity in various scenarios~\cite{ma_sam,samrs,surgicalsam,alignsam}. However, SAM still faces some limitations when applied to downstream medical tasks. As illustrated in Fig.~\ref{fig:title} (a$^{(i)}$), vanilla SAM~\cite{SAM} heavily relies on manual prompts, such as points and boxes. However, due to the small and numerous nature of DR lesions, manual prompts become labor-intensive, rendering such an approach impractical for clinical applications. 
Some methods introduce in-context prompts to adapt SAM from the global perspective, like one-shot (Fig.~\ref{fig:title} (a$^{(ii)}$)) prompt~\cite{persam}, but they struggle to handle local lesions, resulting in suboptimal performance.
Parameter-efficient fine-tuning methods~\cite{arXiv23_ophSAM,mammosam} (Fig.~\ref{fig:title} (a$^{(iii)}$)) adapt SAM for downstream tasks by tuning a limited number of parameters.
However, these methods overlook the prompt-based strategy to achieve automatic inference. A more flexible approach to prompt segmentation would be preferred in practice, allowing for physicians to refine results with greater precision through targeted prompts when necessary.
Moreover, these SAM-related methods struggle to distinguish fine-grained DR lesion categories and typically can only generate class-agnostic masks.

Vision-Language Models (VLMs), with their capability to align images with corresponding textual descriptions, have shown remarkable effectiveness in many downstream applications~\cite{stablediffusion,crowson2022vqgan,xia2023hgclip,xia2024cares,tang2024hunting,xing2024hybrid,xing2024segmamba}. 
This raises a possibility: \textit{Could VLM assist visual models in locating lesions using textual cues, further enhancing the accuracy of distinguishing different lesions and enhancing the credibility of segmentation? }
However, the potential of VLMs remains largely untapped in medical imaging primarily due to the significant gap between natural and medical domains. As shown in Fig.~\ref{fig:title} (b), the class activation maps generated by implicit class name reveal that the VLM (\textit{e.g.}, CLIP~\cite{clip}) fails to provide useful priors in this context.

In this study, we concentrate on designing a flexible scheme for segmenting DR lesions, allowing the direct generation of masks corresponding to specific text-based categories, as illustrated in Fig.~\ref{fig:title} (a$^{(iv)}$). 
Meanwhile, we aim at enhancing the model's credibility and its ability to discern inter-class differences in DR lesions by integrating text-based cues via VLM.
Thus, we propose an explicit prior encoder that utilizes \textit{explicit description of lesions} rather than \textit{implicit class name} to generate explainable cues for segmentation and distinguishing inter-class differences. Specifically, the morphological appearance of DR lesions can be represented by specific descriptions that VLMs easily understand, such as depicting hard exudates as yellowish-white deposits. These explainable cues improve the trustworthiness in the segmentation process.
Further, we introduce a prior-aligned injector into the SAM encoder to inject the text-based external priors into the segmentation process, further facilitating knowledge sharing and alignment between the VLM and vision-only model. 
Lastly, the class-specific prompt generator generates the specific prompts tailored to the text-prompt input, which are subsequently fed into the SAM decoder to produce the corresponding segmentation mask.

Our main contributions are as follows. First, we propose a novel framework that exploits explainable cues generated from textual prompts, thereby enhancing the reliability of DR segmentation. 
Second, we introduce an explicit prior encoder to transfer implicit medical concepts into explicit priors, providing explainable global guidance for segmentation and enhancing the lesion discrimination ability. 
Third, we design a prior-aligned injector that integrates explainable explicit priors into the segmentation process and facilitates knowledge sharing across multiple modalities.

\section{Methodology}

\textbf{Problem Definition.} 
Given an input image $\mathbf{I}\in\mathbb{R}^{C\times H\times W}$ and a text class-prompt $t$ of the $i$-${th}$ class $c_i$, our goal is to generate the mask $\mathbf{M}_i$ of $c_i$.

\begin{figure*}[t]
\includegraphics[width=\textwidth]{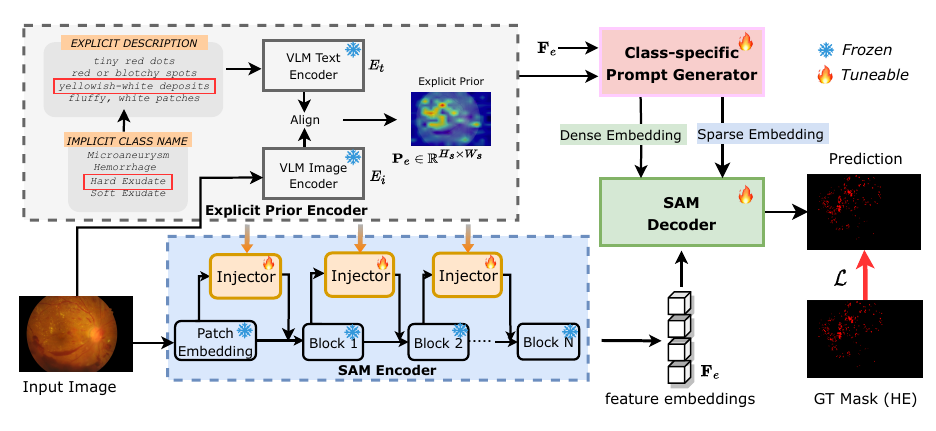}
\caption{
Overview of the proposed framework. The explicit prior encoder extracts explainable clues and generate prior knowledge for segmentation. This explicit prior is then fed in to the prior-aligned injector (Injector) for injecting prior knowledge to feature encoding process. The class-specific prompt generator produces the segmentation map according to the provided text-based class.
} 
\label{fig:main}
\end{figure*}

\noindent
\textbf{Framework Overview.} 
Fig.~\ref{fig:main} illustrates the overview of our method, consisting of four key components: a VLM-based explicit prior encoder, a SAM encoder with the prior-aligned injector, a class-specific prompt generator, and a SAM decoder. 
The explicit prior encoder first encodes the text class-prompt $t$ and produces the explicit prior.
Next, the SAM encoder extracts multi-level features of the input image, and the prior-aligned injectors facilitate knowledge sharing between the text-guided explicit prior and multi-level visual features.
Then, the class-specific prompt generator generates prompts based on explicit prior, subsequently fed into the SAM decoder to produce the corresponding mask.

\subsection{Explicit Prior Encoder} 

Different from the existing vision-only DR segmentation approaches, we resort to the language modality to provide external knowledge in segmentation. We delve into using the \textit{explicit
description} instead of the \textit{implicit class name} to guide segmentation. 
This strategy entails harnessing external knowledge and preprocessing it through the robust image-text knowledge ingrained in the VLM (\textit{e.g.}, CLIP~\cite{clip}), ultimately generating what we term as an \textit{explicit prior}. The incorporation of explicit prior information provides explainable cues that enhance the trustworthiness of the segmentation process.

Specifically, the frozen image encoder $E_i(\cdot)$ and text encoder $E_t(\cdot)$ of pre-trained CLIP are utilized to encode the visual input and the explicit lesion knowledge to get visual prior and text prior as $\mathbf{P}_v=E_i(\mathbf{I})\in\mathbb{R}^{H_s\times W_s\times C_t}$ and $\mathbf{P}_t=E_t(t)\in\mathbb{R}^{1\times C_t}$, where $C_t$ is the dimension of the features. 
Then we reshape $\mathbf{P}_v$ into $\mathbf{P}_v=E_i(\mathbf{I})\in\mathbb{R}^{(H_sW_s)\times C_t}$, and align text prior and visual prior as: 
\begin{equation}\small
\mathbf{S}=\frac{\mathbf{P}_v}{||\mathbf{P}_v||_2}\cdot (\frac{\mathbf{P}_t}{||\mathbf{P}_t||_2})^T,
\label{equ:cross_project}
\end{equation}
where $||\cdot||_2$ is the L2 normalization and $\mathbf{S}\in\mathbb{R}^{(H_sW_s)\times 1}$.
Next, to generate explicit prior map, we reshape $\mathbf{S}$ to $\mathbf{S}'\in\mathbb{R}^{H_s\times W_s}$ and map the $\mathbf{S}'$ into explicit prior $\mathbf{P}_e$ as $\mathbf{P}_e=Norm(\mathbf{S}')$, where $Norm(\cdot)$ denotes the min-max normalization.
Using explicit lesion descriptions instead of implicit class names unleashes the potential of the VLM, offering global guidance for subsequent segmentation steps and enhancing the differentiation of inter-class differences. 

\begin{figure*}[h]
\includegraphics[width=\textwidth]{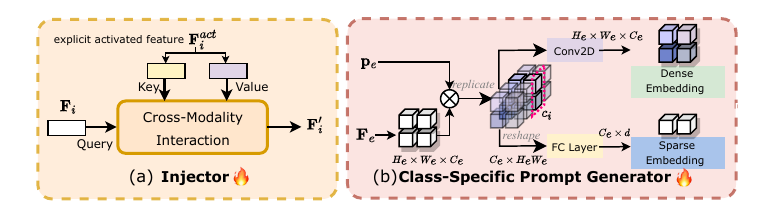}
\caption{
The detailed architectures of (a) prior-aligned injector (Injector)  and (b) class-specific prompt generator.
} 
\label{fig:main2}
\end{figure*}

\subsection{Prior-Aligned Injector}
Since the knowledge embedded in the pre-trained SAM and CLIP models do not "see" each other and remain isolated before integrating, 
it is crucial to construct a bridge for interaction, ensuring the alignment of representations within a unified feature space. 
Moreover, a mechanism is necessary to inject external knowledge into the segmentation process.
To address this, we propose a prior-aligned injector (shown in Fig.~\ref{fig:main2}(a)) in each encoder layer, aiming to facilitate knowledge sharing between the segmentation and vision-language models. 
Formally, for the intermediate $i$-$th$ block in SAM encoder, the encoded feature $\mathbf{F}_i\in\mathbb{R}^{H_s\times W_s\times C_i}$ is fed into the cross-modality interaction module for interacting with explicit prior $\mathbf{P}_e$. 
The output of injector $\mathbf{F}_i'$ is fed into the next encoder block.\\
\noindent
\textbf{Cross-Modality Interaction.} 
We first aggregate the encoded feature $\mathbf{F}_i$ with explicit prior $\mathbf{P}_e$ to obtain the explicit activated feature as $\mathbf{F}^{act}_i=\mathbf{F}_i\times\mathbf{P}_e$. This operation enables the accurate location prior can be utilized in the interaction process. Then, we take $\mathbf{F}_i$ as the query, and the explicit activated feature $\mathbf{F}'_i$ as both key and value. 
The query, key and value are fed into zoomed projected operation to adjust resolutions, achieved by an $1\times1$ convolutional layer with stride-4, which can be written as:
\begin{equation}\small
\mathbf{F}^Q_i = \phi_q(\mathbf{F}^{act}_i), \mathbf{F}^K_i = \phi_k(\mathbf{F}^{act}_i), \mathbf{F}^V_i = \phi_v(\mathbf{F}^{act}_i),
\label{equ:cross_project}
\end{equation}
where $\phi_q(\cdot)$, $\phi_k(\cdot)$, $\phi_v(\cdot)$ denote the zoomed projected operation.
Then, we employ these projected features to conduct cross-modality interaction, as:
\begin{equation}\small
\mathbf{F}_i'=\gamma \mathbf{F}_i + softmax(\frac{\mathbf{F}^Q_i\cdot{\mathbf{F}^K_i}^T}{\sqrt{d}}){\mathbf{F}^V_i},
\label{equ:cross}
\end{equation}
where $d$ is the dimension of the key vectors, $\gamma$ is a learnable parameter used to adjust the blending ratio of the attention output with the original input. 
In this module, we introduce residual connections to enhance the stability.
Finally, we adjust the resolution of $\mathbf{F}_i'$ back to the input size through upsampling, as $\mathbf{F}_i''=Upsample(\mathbf{F}_i')$. 
By doing so, this injector can model the contexts of images with the guidance of textual global prior without fully fine-tuning the encoder. 

\subsection{Class-specific Prompt Generator}
The explicit prior based on text prompt is included in this module to guide the prompt generation for lesion segmentation, as shown in Fig.~\ref{fig:main2} (b).
The feature embeddings generated by the image encoder are denoted as $\mathbf{F}_e\in\mathbb{R}^{H_e W_e \times C_e}$, where $H_e, W_e$ and $C_e$ denote the width, height, and channel of the feature, respectively. We reshape it as $\mathbf{F}_e\in\mathbb{R}^{H_e  \times W_e \times C_e}$. Then, it interacts with the prior feature $\mathbf{P}_e$ to get the prior-guided feature as $\mathbf{F}_p=\mathbf{F}_e\times\mathbf{P}_e$. We replicate $\mathbf{F}_p$ $c$ times to get $\mathbf{F}'_p\in\mathbb{R}^{c\times H_e  \times W_e \times C_e}$ and assign a specific channel for each category, where $c$ is the overall class numbers. By doing so, each channel contains category-specific information. For the given class $c_i$, we keep only the channels relevant to the given class $c_i$ (as $\mathbf{F}^{c_i}_p$) and project it to get dense embeddings $\mathbf{E}_d$ and sparse embeddings $\mathbf{E}_s$ as 
\begin{equation}\small
\mathbf{E}_d=\phi_{dense}(\mathbf{F}^{c_i}_p), \mathbf{E}_s=\phi_{sparse}(reshape(\mathbf{F}^{c_i}_p)),
\label{equ:cross}
\end{equation}
where $\phi_{dense}(\cdot)$ is the convolution operation, $\phi_{sparse}(\cdot)$ is the linear projection operation, and $reshape(\cdot)$ operation reshapes the feature as $\mathbf{F}^{c_i}_p\in\mathbb{R}^{C_e\times H_e W_e}$. Subsequently, $\mathbf{E}_d$ and $\mathbf{E}_s$ are fed into the SAM decoder, serving as inputs for the dense and sparse embeddings within the original SAM decoder. 
Here, we leverage the original SAM decoder to process dense and sparse embeddings for lesion segmentation. Dense embeddings offer global guidance, while sparse embeddings retain more detailed information about the lesions, further boosting lesion segmentation.
Finally, the SAM decoder outputs the prediction map $\mathbf{P}$.

\noindent
\textbf{Training Objective.}
To train our segmentation model, the overall training objective adopts the combination of binary cross entropy loss and IoU loss, which is defined as $\mathcal{L}=\frac{1}{N}\sum_{j=1}^N(\mathcal{L}_{IoU}(\mathbf{P}_j,\mathbf{G}_j)+\mathcal{L}_{BCE}(\mathbf{P}_j,\mathbf{G}_j)),$
where $\mathbf{P}_j$ is the prediction map and $\mathbf{G}_j$ is the ground truth map.

\section{Experiments}
\subsection{Experimental Setup}
\textbf{Dataset.}
To evaluate the effectiveness of our method, we adopt IDRiD~\cite{idrid} and DDR~\cite{DDR} datasets, which both contain four categories (MA, HE, EX, and SE) of DR lesion region from color fundus images.

\noindent
\textbf{Implementation Details.}
Our method is implemented with Pytorch and all experiments are conducted on 2$\times$NVIDIA RTX 4090 GPUs. We train the IDRiD dataset for 500 epochs and train the DDR dataset for 200 epochs.
We resize the input images to $1024\times1024$ in the training and inference stages.
The AdamW optimizer is adopted with a learning rate of 0.0001. For SAM and CLIP, we use their ViT-B~\cite{vit} and ViT-B/16 variants, respectively. The explicit descriptions are crafted from ophthalmology literature and have undergone validation by both relevant experts and GPT-4. 

\subsection{Comparison Study}
Firstly, we compare our method with the SOTA specialized segmentation model, including U-Net~\cite{unet}, Fully Convolutional Transformer (FCT)~\cite{fct}. 
Secondly, we compare our method with SAM~\cite{SAM} and its variants, including MedSAM~\cite{ma_sam}, SAMed~\cite{samed}, PerSAM/PerSAM-F~\cite{persam} and SAM-Adapter~\cite{sam_adapter}. 

\begin{table}[t!]\small
\caption{Model performance on the IDRiD~\cite{idrid} and DDR~\cite{DDR} datasets. 
} 
\centering
\setlength{\tabcolsep}{2.4mm}{
\scalebox{0.89}{
\begin{tabular}{lcccccc} 
\Xhline{0.7pt}
Dataset & \multicolumn{3}{c}{IDRiD~\cite{idrid}} & \multicolumn{3}{c}{DDR~\cite{DDR}} \\
\cmidrule(r){1-1} \cmidrule(r){2-4} \cmidrule(r){5-7}
Method & mDice & AUC-ROC & AUC-PR  & mDice & AUC-ROC & AUC-PR \\ 
\cmidrule(r){1-1} \cmidrule(r){2-4} \cmidrule(r){5-7}
U-Net~\cite{unet}      &  36.66  & 89.34 & 34.68 & 24.59 & 91.36 & 26.78\\
FCT~\cite{fct}  &  42.96  & 93.49 & 43.09 & 23.33 & 90.53 & 25.01 \\

\cmidrule(r){1-1} \cmidrule(r){2-4} \cmidrule(r){5-7}
Vanilla SAM~\cite{SAM}   & 1.35 & 45.35 & 0.61 & 0.51  & 34.56 & 0.28\\
PerSAM~\cite{persam}           & 1.65 & 61.20 & 0.94 & 0.76 & 69.04 & 0.59\\
PerSAM-F~\cite{persam}        & 1.64 & 57.98 & 0.73 & 0.72 & 56.41 & 0.57\\
MedSAM~\cite{ma_sam}   & 25.06 & 87.27 & 26.57 & 17.97 & 89.83 & 19.30\\
SAMed~\cite{samed}   & 35.18 & 88.96 & 35.37 & 28.32 & 87.70 & 30.03\\
SAM-Adapter~\cite{sam_adapter} & 34.42 & 88.23 & 34.59 & 29.11 & 88.82 & 29.90\\
\cmidrule(r){1-1} \cmidrule(r){2-4} \cmidrule(r){5-7}
\textbf{Ours}       & \textbf{49.72} & \textbf{95.94} & \textbf{50.55} & \textbf{38.78} & \textbf{94.12} & \textbf{39.15} \\
\Xhline{0.7pt}
\end{tabular}}
}
\label{compare1}
\end{table}
\begin{table}\small
\caption{Model performance under different parameter-efficient fine-tuning strategies. 
} 
\centering
\setlength{\tabcolsep}{2mm}{
\scalebox{0.85}{
\begin{tabular}{lcccccc} 
\Xhline{0.7pt}
Dataset & \multicolumn{3}{c}{IDRiD~\cite{idrid}} & \multicolumn{3}{c}{DDR~\cite{DDR}} \\
\cmidrule(r){1-1} \cmidrule(r){2-4} \cmidrule(r){5-7}
Method & mDice & AUC-ROC & AUC-PR  & mDice & AUC-ROC & AUC-PR \\ 
\cmidrule(r){1-1} \cmidrule(r){2-4} \cmidrule(r){5-7}
Adapter~\cite{MLP_fintune}          & 47.46 & 95.55 & 49.67 & 34.20 & \textbf{95.71} & 35.99\\
Multi-scale Adapter~\cite{mammosam} & 45.95 & 95.16 & 47.72 & 35.42 & 94.58 & 36.81\\
LoRA~\cite{lora}  & 44.22 & 95.07 & 47.28 & 36.77 & 94.51 & 37.58\\
\textbf{Prior-Aligned Injector}       & \textbf{49.72} & \textbf{95.94} & \textbf{50.55} &\textbf{38.78} & 94.12 & \textbf{39.15} \\
\Xhline{0.7pt}
\end{tabular}}
}
\label{compare_adapter}
\end{table}

The experimental results in Table~\ref{compare1} show that our TP-DRSeg outperforms other competitive methods on both IDRiD and DDR datasets.
Using zero-shot (vanilla) SAM can not achieve satisfactory performance due to the gap between medical and natural domains. Meanwhile, in-context prompted SAM~\cite{persam} is unable to identify fine-grained features and generate accurate prompts, consequently yielding suboptimal segmentation results.
Those SAM variants~\cite{ma_sam,samed,sam_adapter} improve performance with parameter-efficient fine-tuning but only use visual cues, limiting their ability to exploit clues from other modalities. Our approach, integrating visual and language cues, demonstrates better segmentation performance.

\begin{figure}[t]
    \centering
    \includegraphics[width=1.0\linewidth]{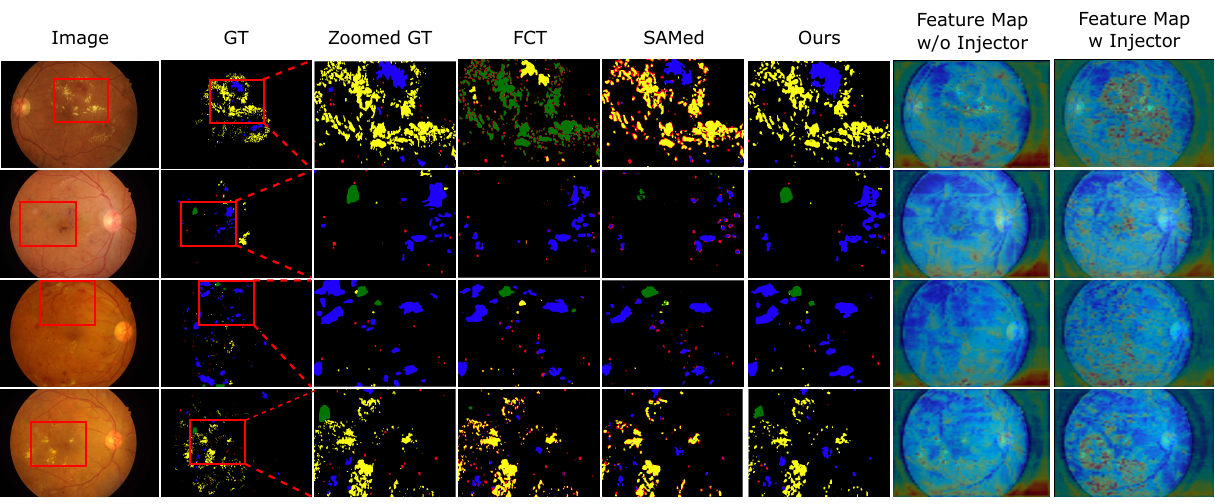}
    \caption{Qualitative comparison and visualized feature maps with and without the integration of explicit prior. }
    \label{fig:compare}
\end{figure}

\noindent
\textbf{Qualitative Analysis.} The qualitative results are shown in Fig.~\ref{fig:compare} and we can see that our method exhibits greater advantages in lesion identification. 
FCT can locate lesion regions but struggles with distinguishing HE from SE, showing the value of textual information in distinguishing inter-class differences.
Additionally, we show the visualized features with and without the integration of explicit priors. It is evident that text-based explicit priors can serve as explainable clues for segmentation and can guide the network in recognizing lesions. 

\begin{table}[ht]\small
\caption{Ablation analysis of the components in our framework.} 
\centering
\setlength{\tabcolsep}{2.1mm}{
\scalebox{0.85}{
\begin{tabular}{lcccccc} 
\Xhline{0.7pt}
Dataset & \multicolumn{3}{c}{IDRiD~\cite{idrid}} & \multicolumn{3}{c}{DDR~\cite{DDR}} \\
\cmidrule(r){1-1} \cmidrule(r){2-4} \cmidrule(r){5-7}
Method & mDice & AUC-ROC &  AUC-PR & mDice & AUC-ROC & AUC-PR \\ 
\cmidrule(r){1-1} \cmidrule(r){2-4} \cmidrule(r){5-7}
Ours-w/o Injector        & 35.09 & 90.35 & 33.64 & 24.07 & 91.99 & 22.74\\
Ours-w/o EP in Injector  & 44.89 & 94.93 & 46.61 & 37.27 & 93.99 & 38.47\\
Ours-w/o EP in CPG       & 45.03 & 94.65 & 47.21 & 35.25 & 92.82 & 36.65\\
Ours                     & \textbf{49.72} & \textbf{95.94} & \textbf{50.55} & \textbf{38.78} & \textbf{94.12} & \textbf{39.15} \\
\Xhline{0.7pt}
\end{tabular}}
}
\label{abl_class}
\end{table}
\noindent
\textbf{Comparison with Other Parameter-Efficient Fine-Tuning Methods.} Moreover, we compare our prior-aligned injector with other parameter-efficient fine-tuning methods, including Adapter~\cite{MLP_fintune}, LoRA~\cite{lora} and Multi-Scale Adapter~\cite{mammosam}, which is shown in Table~\ref{compare_adapter}. The compared methods target at single modality, \textit{i.e.} vision modality, without considering the text-guided information. Our proposed prior-aligned injector leverages explicit textual guidance to help locate DR lesions and demonstrate better segmentation capability.

\subsection{Ablation Study}
The results of the ablation study are shown in Table~\ref{abl_class}. We can observe the performance degradation when removing the prior-aligned injector (Injector) in training. Then we test the model performance with and without the integration of explicit prior (EP) in the Injector and in the class-specific prompt generator (CPG).
The experimental results in the table demonstrate that the text-based explicit prior is effective in boosting lesion segmentation.

\section{Conclusion}
In this paper, we focus on how language cues benefit DR lesion segmentation and propose a novel framework in a text-prompted scheme, termed TP-DRSeg.
Specifically, we design an explicit prior encoder to provide explainable clues with text-based prompts.
We also introduce a prior-aligned injector to efficiently inject explicit prior knowledge in the segmentation process and enable our framework training in a parameter-efficient fashion.
The experimental results demonstrate the superiority and effectiveness of our proposed TP-DRSeg.
%
%

\begin{credits}
\subsubsection{\ackname}
The work was partially supported by Airdoc medical AI projects donation Phase 2, Monash-Airdoc Research Centre, and in part by the MRFF NCRI GA89126.

\subsubsection{\discintname}
The authors declare that they have no competing interests.
\end{credits}

\bibliographystyle{splncs04}
\bibliography{ref}

\begin{thebibliography}{10}
\providecommand{\url}[1]{\texttt{#1}}
\providecommand{\urlprefix}{URL }
\providecommand{\doi}[1]{https://doi.org/#1}

\bibitem{sam_adapter}
Chen, T., Zhu, L., Ding, C., Cao, R., Zhang, S., Wang, Y., Li, Z., Sun, L., Mao, P., Zang, Y.: Sam fails to segment anything?--sam-adapter: Adapting sam in underperformed scenes: Camouflage, shadow, and more. arXiv:2304.09148  (2023)

\bibitem{crowson2022vqgan}
Crowson, K., Biderman, S., Kornis, D., Stander, D., Hallahan, E., Castricato, L., Raff, E.: Vqgan-clip: Open domain image generation and editing with natural language guidance. In: European Conference on Computer Vision. pp. 88--105. Springer (2022)

\bibitem{vit}
Dosovitskiy, A., Beyer, L., Kolesnikov, A., Weissenborn, D., Zhai, X., Unterthiner, T., Dehghani, M., Minderer, M., Heigold, G., Gelly, S., et~al.: An image is worth 16x16 words: Transformers for image recognition at scale. In: International Conference on Learning Representations (2020)

\bibitem{NC19_LSEG}
Guo, S., Li, T., Kang, H., Li, N., Zhang, Y., Wang, K.: L-seg: An end-to-end unified framework for multi-lesion segmentation of fundus images. Neurocomputing  \textbf{349},  52--63 (2019)

\bibitem{BSPC24_MSF}
Guo, T., Yang, J., Yu, Q.: Diabetic retinopathy lesion segmentation using deep multi-scale framework. Biomedical Signal Processing and Control  \textbf{88},  105050 (2024)

\bibitem{CIS22_CARNet}
Guo, Y., Peng, Y.: Carnet: Cascade attentive refinenet for multi-lesion segmentation of diabetic retinopathy images. Complex \& Intelligent Systems  \textbf{8}(2),  1681--1701 (2022)

\bibitem{MLP_fintune}
Houlsby, N., Giurgiu, A., Jastrzebski, S., Morrone, B., De~Laroussilhe, Q., Gesmundo, A., Attariyan, M., Gelly, S.: Parameter-efficient transfer learning for nlp. In: International Conference on Machine Learning. pp. 2790--2799. PMLR (2019)

\bibitem{lora}
Hu, E.J., Wallis, P., Allen-Zhu, Z., Li, Y., Wang, S., Wang, L., Chen, W., et~al.: Lora: Low-rank adaptation of large language models. In: International Conference on Learning Representations (2022)

\bibitem{hu2024ophnet}
Hu, M., Xia, P., Wang, L., Yan, S., Tang, F., Xu, Z., Luo, Y., Song, K., Leitner, J., Cheng, X., Cheng, J., Liu, C., Zhou, K., Ge, Z.: Ophnet: A large-scale video benchmark for ophthalmic surgical workflow understanding (2024)

\bibitem{alignsam}
Huang, D., Xiong, X., Ma, J., Li, J., Jie, Z., Ma, L., Li, G.: Alignsam: Aligning segment anything model to open context via reinforcement learning. In: Computer Vision and Pattern Recognition. pp. 3205--3215 (2024)

\bibitem{SAM}
Kirillov, A., Mintun, E., Ravi, N., Mao, H., Rolland, C., Gustafson, L., Xiao, T., Whitehead, S., Berg, A.C., Lo, W.Y., Dollar, P., Girshick, R.: Segment anything. In: International Conference on Computer Vision. pp. 4015--4026 (2023)

\bibitem{DDR}
Li, T., Gao, Y., Wang, K., Guo, S., Liu, H., Kang, H.: Diagnostic assessment of deep learning algorithms for diabetic retinopathy screening. Information Sciences  \textbf{501},  511--522 (2019)

\bibitem{ma_sam}
Ma, J., He, Y., Li, F., Han, L., You, C., Wang, B.: Segment anything in medical images. Nature Communications  \textbf{15}(1), ~654 (2024)

\bibitem{idrid}
Porwal, P., Pachade, S., Kamble, R., Kokare, M., Deshmukh, G., Sahasrabuddhe, V., Meriaudeau, F.: Indian diabetic retinopathy image dataset (idrid): a database for diabetic retinopathy screening research. Data  \textbf{3}(3), ~25 (2018)

\bibitem{arXiv23_ophSAM}
Qiu, Z., Hu, Y., Li, H., Liu, J.: Learnable ophthalmology sam. arXiv preprint arXiv:2304.13425  (2023)

\bibitem{clip}
Radford, A., Kim, J.W., Hallacy, C., Ramesh, A., Goh, G., Agarwal, S., Sastry, G., Askell, A., Mishkin, P., Clark, J., et~al.: Learning transferable visual models from natural language supervision. In: International Conference on Machine Learning. pp. 8748--8763. PMLR (2021)

\bibitem{stablediffusion}
Rombach, R., Blattmann, A., Lorenz, D., Esser, P., Ommer, B.: High-resolution image synthesis with latent diffusion models. In: Computer Vision and Pattern Recognition. pp. 10684--10695 (2022)

\bibitem{unet}
Ronneberger, O., Fischer, P., Brox, T.: U-net: Convolutional networks for biomedical image segmentation. In: International Conference on Medical Image Computing and Computer-Assisted Intervention. pp. 234--241. Springer (2015)

\bibitem{tang2024hunting}
Tang, F., Xu, Z., Qu, Z., Feng, W., Jiang, X., Ge, Z.: Hunting attributes: Context prototype-aware learning for weakly supervised semantic segmentation. In: Computer Vision and Pattern Recognition. pp. 3324--3334 (2024)

\bibitem{fct}
Tragakis, A., Kaul, C., Murray-Smith, R., Husmeier, D.: The fully convolutional transformer for medical image segmentation. In: Winter Conference on Applications of Computer Vision. pp. 3660--3669 (2023)

\bibitem{samrs}
Wang, D., Zhang, J., Du, B., Xu, M., Liu, L., Tao, D., Zhang, L.: Samrs: Scaling-up remote sensing segmentation dataset with segment anything model. Advances in Neural Information Processing Systems  \textbf{36} (2024)

\bibitem{NC23_CLCNet}
Wang, X., Fang, Y., Yang, S., Zhu, D., Wang, M., Zhang, J., Zhang, J., Cheng, J., Tong, K.y., Han, X.: Clc-net: Contextual and local collaborative network for lesion segmentation in diabetic retinopathy images. Neurocomputing  \textbf{527},  100--109 (2023)

\bibitem{xia2024cares}
Xia, P., Chen, Z., Tian, J., Gong, Y., Hou, R., Xu, Y., Wu, Z., Fan, Z., Zhou, Y., Zhu, K., et~al.: Cares: A comprehensive benchmark of trustworthiness in medical vision language models. arXiv preprint arXiv:2406.06007  (2024)

\bibitem{xia2024generalizing}
Xia, P., Hu, M., Tang, F., Li, W., Zheng, W., Ju, L., Duan, P., Yao, H., Ge, Z.: Generalizing to unseen domains in diabetic retinopathy with disentangled representations. arXiv preprint arXiv:2406.06384  (2024)

\bibitem{xia2023hgclip}
Xia, P., Yu, X., Hu, M., Ju, L., Wang, Z., Duan, P., Ge, Z.: Hgclip: Exploring vision-language models with graph representations for hierarchical understanding. arXiv preprint arXiv:2311.14064  (2023)

\bibitem{xie2021survey}
Xie, X., Niu, J., Liu, X., Chen, Z., Tang, S., Yu, S.: A survey on incorporating domain knowledge into deep learning for medical image analysis. Medical Image Analysis  \textbf{69},  101985 (2021)

\bibitem{xing2024segmamba}
Xing, Z., Ye, T., Yang, Y., Liu, G., Zhu, L.: Segmamba: Long-range sequential modeling mamba for 3d medical image segmentation. arXiv preprint arXiv:2401.13560  (2024)

\bibitem{xing2024hybrid}
Xing, Z., Zhu, L., Yu, L., Xing, Z., Wan, L.: Hybrid masked image modeling for 3d medical image segmentation. IEEE Journal of Biomedical and Health Informatics  (2024)

\bibitem{mammosam}
Xiong, X., Wang, C., Li, W., Li, G.: Mammo-sam: Adapting foundation segment anything model for automatic breast mass segmentation in whole mammograms. In: International Workshop on Machine Learning in Medical Imaging. pp. 176--185. Springer (2023)

\bibitem{surgicalsam}
Yue, W., Zhang, J., Hu, K., Xia, Y., Luo, J., Wang, Z.: Surgicalsam: Efficient class promptable surgical instrument segmentation. arXiv preprint arXiv:2308.08746  (2023)

\bibitem{samed}
Zhang, K., Liu, D.: Customized segment anything model for medical image segmentation. arXiv preprint arXiv:2304.13785  (2023)

\bibitem{persam}
Zhang, R., Jiang, Z., Guo, Z., Yan, S., Pan, J., Dong, H., Gao, P., Li, H.: Personalize segment anything model with one shot. In: International Conference on Learning Representations (2024)

\end{thebibliography}

\clearpage
\setcounter{page}{1}


\end{document}